\documentclass{article} 
\PassOptionsToPackage{sort}{natbib}
\usepackage{geometry}
\usepackage[sc]{mathpazo}
\usepackage{microtype}
\usepackage[utf8]{inputenc} 
\usepackage[T1]{fontenc}   
\usepackage{fullpage}
\usepackage[numbers]{natbib}
\usepackage{booktabs}       
\usepackage{amsfonts}       
\usepackage{nicefrac}       
\usepackage{microtype}      
\usepackage{wrapfig,lipsum,booktabs}
\usepackage{multicol}
\usepackage{enumitem}
\usepackage{graphicx}
\usepackage{multicol}
\usepackage{amssymb}
\usepackage{algorithm}
\usepackage{algorithmic}
\usepackage{amsmath}
\usepackage{amsthm}
\usepackage{xcolor}
\usepackage{subcaption}
\usepackage{hyperref}
\usepackage{wrapfig}
\usepackage{xspace} 
\usepackage{enumitem}
\usepackage{caption}
\usepackage{nameref}

\newcommand{\fedavg}{\texttt{FedAvg}\xspace}
\newcommand{\fedprox}{\texttt{FedProx}\xspace}
\newcommand{\mocha}{\texttt{MOCHA}\xspace}

\title{Federated Learning: \\ Challenges, Methods, and Future Directions}

\author{\begin{tabular}{cc}
\small Tian Li & \small Anit Kumar Sahu \tabularnewline 
\small Carnegie Mellon University & \small Bosch Center for Artificial Intelligence \tabularnewline
\small{\texttt{\href{mailto:tianli@cmu.edu}{tianli@cmu.edu}}} & \small{\texttt{\href{mailto:anit.sahu@gmail.com}{anit.sahu@gmail.com}}} \vspace{3mm} \tabularnewline 
\small Ameet Talwalkar & \small Virginia Smith \tabularnewline 
\small Carnegie Mellon University \& Determined AI & \small Carnegie Mellon University \tabularnewline
\small{\texttt{\href{mailto:talwalkar@cmu.edu}{talwalkar@cmu.edu}}} & \small{\texttt{\href{mailto:smithv@cmu.edu}{smithv@cmu.edu}}}
\end{tabular}}

\date{}

\begin{document}

\maketitle

\begin{abstract}

Federated learning involves training statistical models over remote devices or siloed data centers, such as mobile phones or hospitals, while keeping data localized. Training in heterogeneous and potentially massive networks introduces novel challenges that require a fundamental departure from standard approaches for large-scale machine learning, distributed optimization, and privacy-preserving data analysis. In this article, we discuss the unique characteristics and challenges of federated learning, provide a broad overview of current approaches, and outline several directions of future work that are relevant to a wide range of research communities. 
\end{abstract}

\section{Introduction}
\label{sec: intro}

Mobile phones, wearable devices, and autonomous vehicles are just a few of the modern distributed networks generating a wealth of data each day.
Due to the growing computational power of these devices---coupled with concerns over transmitting private information---it is increasingly attractive to store data \emph{locally} and push  network computation to the edge. 

The concept of edge computing is not a new one.  Indeed, computing simple queries across distributed, low-powered
devices is a decades-long area of research that has been explored under the purview of query processing
in sensor networks, computing at the edge, and fog computing~\cite{madden2005tinydb,deshpande2005model, bonomi2012fog, hong2013mobile, garcia2015edge}. Recent works have also considered training machine learning models centrally but serving and storing them locally; for example, this is a common approach in mobile user modeling and personalization~\cite{kuflik2012challenges, rastegari2016xnor}. 

\begin{figure}[ht!]
    \centering
    \includegraphics[width=0.8\textwidth]{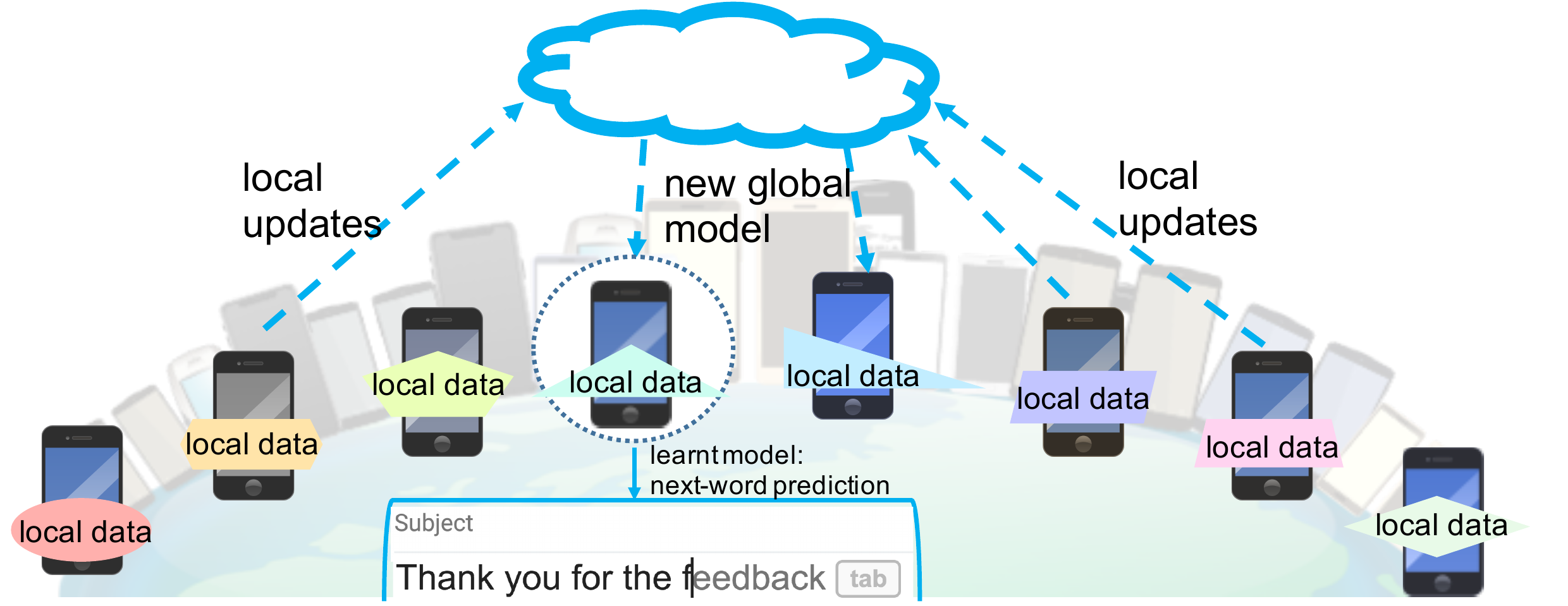}
    \caption{An example application of federated learning for the task of next-word prediction on mobile phones. To preserve the privacy of the text data and to reduce strain on the network, we seek to train a predictor in a distributed fashion, rather than sending the raw data to a central server. In this setup, remote devices communicate with a central server periodically to learn a global model. At each communication round, a subset of selected phones performs local training on their non-identically-distributed user data, and sends these local updates to the server. After incorporating the updates, the server then sends back the new global model to another subset of devices. This iterative training process continues across the network until convergence is reached or some stopping criterion is met.} 
    \label{fig: overview}
\end{figure}

However, as the storage and computational capabilities of the devices within distributed networks grow, it is possible to leverage enhanced local resources on each device. This  has led to a growing interest in \emph{federated learning}~\cite{mcmahan2016FedAvg}, which
explores \emph{training} statistical models directly on remote devices\footnote{We use the term `device' throughout the article to describe entities in the network, such as nodes, clients, sensors, or organizations.}.
As we discuss in this article, learning in such a setting differs significantly from traditional distributed environments---requiring fundamental advances in areas such as privacy, large-scale machine learning, and distributed optimization, and raising new questions at the intersection of diverse fields, such as machine learning and systems~\cite{sysml-white}.

Federated learning methods have been deployed by major service providers~\cite{bonawitz2019towards,fedai}, and play a critical role in supporting privacy-sensitive applications where the training data are distributed at the edge~\citep[e.g.,][]{zhao2019mobile, ramaswamy2019federated, huang2019patient, silva2018federated, hard2018federated, yang2019federated, ammad2019federated}. 
Examples of potential applications include: learning sentiment, semantic location, or activities of mobile phone users; adapting to pedestrian behavior in autonomous vehicles; and predicting health events like heart attack risk from wearable devices~\cite{anguita2013public,pantelopoulos2010survey, huang2018loadaboost}. 
We discuss several canonical applications of federated learning below:

\begin{itemize}

\vspace{-1em}
    \item \emph{Smart phones.} By jointly learning user behavior across a large pool of mobile phones, statistical models can power applications such as next-word prediction, face detection, and voice recognition~\cite{ramaswamy2019federated,hard2018federated}.
    However, users may not be willing to share their data in order to protect their personal privacy or to save the limited bandwidth/battery power of their phone.
    Federated learning has the potential to enable predictive features on smart phones without diminishing the user experience or leaking private information. Figure \ref{fig: overview} depicts one such application in which we aim to learn a next-word predictor in a large-scale mobile phone network based on users' historical text data~\cite{hard2018federated}. 
    \item \emph{Organizations.} Organizations or institutions can also be viewed as `devices' in the context of federated learning. 
    For example, hospitals are organizations that contain a multitude of patient data for predictive healthcare. However, hospitals operate under strict privacy practices, and may face legal, administrative, or ethical
    constraints that require data to remain local. Federated learning is a promising solution for these applications~\cite{huang2018loadaboost}, as it can reduce strain on the network and enable private learning between various devices/organizations.
    
    \item \emph{Internet of things.} 
    Modern IoT networks, such as wearable devices, autonomous vehicles, or smart homes, may  contain numerous sensors that allow them to collect, react, and adapt to incoming data in real-time. For example, a fleet of autonomous vehicles may require an up-to-date model of traffic, construction, or pedestrian behavior to safely operate. However, building aggregate models in these scenarios may be difficult due to the private nature of the data and the limited connectivity of each device. Federated learning methods can help to train models that efficiently adapt to changes in these systems while maintaining user privacy~\cite{Samarakoon2018globecom,pantelopoulos2010survey}.

\end{itemize}

\subsection{Problem Formulation}

The canonical federated learning problem involves learning a \emph{single, global} statistical model from data stored on tens to potentially millions of remote devices. We aim to learn this model under the constraint that 
device-generated data is stored and processed locally, with only intermediate updates being communicated periodically with a central server. In particular, the goal is typically to minimize the following objective function:
\begin{align}
    \min_w \, F(w) \, , \,\,\, \text{where} \,\,\, F(w) := \sum_{k=1}^m p_k F_k(w) \, . \label{eq:original_obj}
\end{align}
Here, $m$ is the total number of devices, $p_k \geq 0$ and $\sum_k p_k=1$, and $F_k$ is the local objective function for the $k$th device. The local objective function is often defined as the empirical risk over local data, i.e., $F_k(w) = \frac{1}{n_k}\sum_{j_k=1}^{n_k} f_{j_k}(w; x_{j_k}, y_{j_k})$, where $n_k$ is the number of samples available locally. The user-defined term $p_k$ specifies the relative impact of each device, with two natural settings being  $p_k=\frac{1}{n}$ or $p_k=\frac{n_k}{n}$, where $n=\sum_k n_k$ is the total number of samples. We will reference problem~\eqref{eq:original_obj} throughout the article, but, as  discussed below, we note that other objectives or modeling approaches may be appropriate depending on the application of interest. 

\subsection{Core Challenges}
We next describe four of the core challenges associated with solving the distributed optimization problem posed in \eqref{eq:original_obj}. These challenges make the federated setting distinct from other classical problems, such as distributed learning in data center settings or traditional private data analyses.

{\bf Challenge 1: Expensive Communication.} 
Communication is a critical bottleneck in federated networks, which, coupled with privacy concerns over sending raw data, necessitates that data generated on each device remain local.
Indeed, federated networks are potentially comprised of a massive number of devices, e.g., millions of smart phones, and communication in the network can be slower than local computation by many orders of magnitude~\cite{van2009multi,huang2013depth}. 
In order to fit a model to data generated by the devices in the federated network, it is therefore necessary to develop communication-efficient methods that iteratively send small messages or \textit{model updates} as part of the training process, as opposed to sending the entire dataset over the network. To further reduce communication in such a setting, two key aspects to consider are: (i) reducing the total number of communication rounds, or (ii) reducing the size of transmitted messages at each round. 

{\bf Challenge 2: Systems Heterogeneity.} The storage, computational, and communication capabilities of each device in federated networks may differ due to variability in hardware (CPU, memory), network connectivity (3G, 4G, 5G, wifi), and power (battery level). Additionally, the network size and systems-related constraints on each device typically result in only a small fraction of the devices being active at once, e.g., hundreds of active devices in a million-device network~\cite{bonawitz2019towards}. Each device may also be unreliable, and it is not uncommon for an active device to drop out at a given iteration due to connectivity or energy constraints.  These system-level characteristics dramatically exacerbate challenges such as straggler mitigation and fault tolerance. Federated learning methods that are developed and analyzed must therefore: (i) anticipate a low amount of participation, (ii) tolerate heterogeneous hardware, and (iii) be robust to dropped devices in the network.

{\bf Challenge 3: Statistical Heterogeneity.} Devices frequently generate and collect data in a non-identically distributed manner across the network, e.g.,  mobile phone users have varied use of language in the context of a next word prediction task. Moreover, the number of data points across devices may  vary significantly, and there may be an underlying structure present that captures the relationship amongst devices and their associated distributions. 
This data generation paradigm violates frequently-used independent and identically distributed~(I.I.D.) assumptions in distributed optimization, increases the likelihood of stragglers, and may add complexity in terms of modeling, analysis, and evaluation. 
Indeed, although the canonical federated learning problem of \eqref{eq:original_obj} aims to learn a single global model,
there exist other alternatives such as simultaneously learning distinct local models 
via multi-task learning frameworks~\cite[cf.][]{fed_multitask_smith_2017}. There is also a close connection in this regard between leading approaches for federated learning and meta-learning~\cite{li_gbml19}. Both the multi-task and meta-learning perspectives enable \emph{personalized} or \emph{device-specific} modeling, which is often a more natural approach to handle the statistical heterogeneity of the data.

{\bf Challenge 4: Privacy Concerns.} Finally, privacy is often a major concern in federated learning applications. Federated learning makes a step towards protecting data generated on each device by sharing model updates, e.g., gradient information, instead of the raw data~\cite{duchi2014privacy, dwork2014algorithmic,carlini2018secret}. However, communicating model updates throughout the training process can nonetheless reveal sensitive information, either to a third-party, or to the central server~\cite{mcmahan2018diff}. While recent methods aim to enhance the privacy of federated learning using tools such as secure multiparty computation or differential privacy, these approaches often provide privacy at the cost of reduced model performance or system efficiency. Understanding and balancing these trade-offs, both theoretically and empirically, is a considerable challenge in realizing private federated learning systems. 

The remainder of this article is organized as follows. In Section~\ref{sec: related_work}, we introduce previous and current works that aim to  address the four discussed challenges of federated learning. In Section~\ref{sec: future_work}, we outline several promising directions of future research.

\section{Survey of Related and Current Work}
\label{sec: related_work}

The challenges in federated learning at first glance resemble classical problems in areas such as privacy, large-scale machine learning, and distributed optimization. For instance, numerous methods have been proposed to tackle expensive communication in the machine learning, optimization, and signal processing communities. However, these methods are typically unable to fully handle the scale of federated networks, much less the challenges of systems and statistical heterogeneity. 
Similarly, while privacy is an important aspect for many machine learning applications, privacy-preserving methods for federated learning can be challenging to rigorously assert due to the statistical variation in the data, and may be even more difficult to implement due to systems constraints on each device and across the potentially massive network. 
In this section, we explore in more detail the challenges presented in Section~\ref{sec: intro}, including a discussion of classical results 
as well as more recent work focused specifically on federated learning.  

\subsection{Communication-efficiency}
\label{sec: communication}
Communication is a key bottleneck to consider when developing methods for federated networks. While it is beyond the scope of this article to provide a self-contained review of communication-efficient distributed learning methods, we point out several general directions, which we group into (1) local updating methods, (2) compression schemes, and (3) decentralized training.

\subsubsection{Local Updating} 

\begin{figure}[t]
    \centering
    \begin{subfigure}{0.55\textwidth}
    \includegraphics[width=\textwidth]{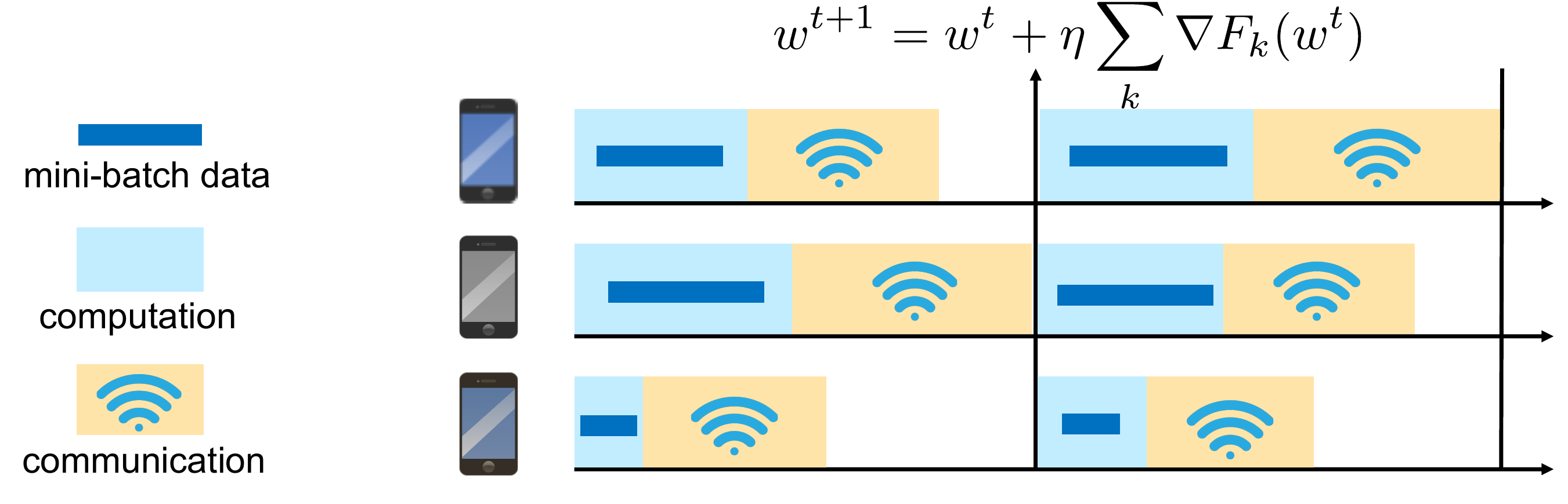}
    \end{subfigure}
    \hspace{0.05\textwidth}
    \begin{subfigure}{0.38\textwidth}
    \includegraphics[width=\textwidth]{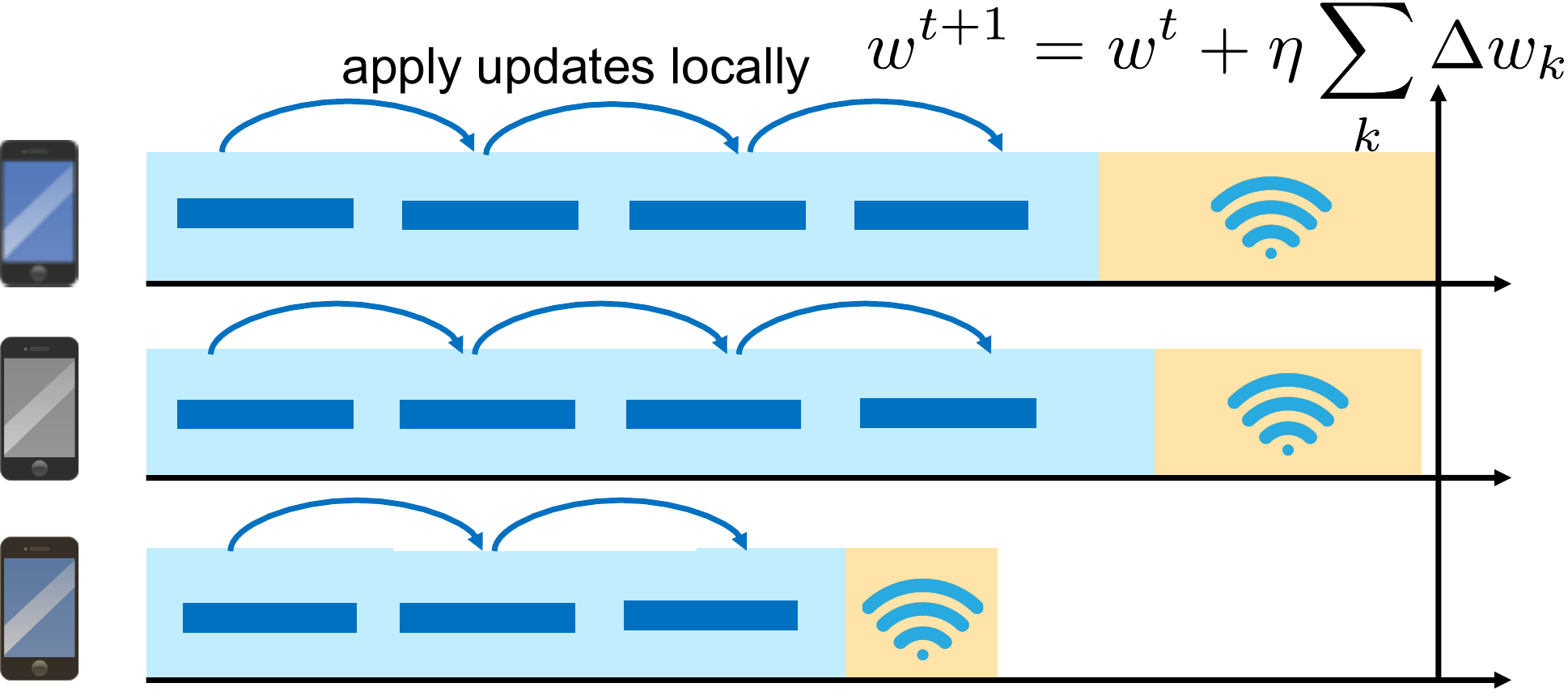}
    \end{subfigure}
    \caption{\textit{Left: Distributed (mini-batch) SGD.} Each device, $k$, locally computes gradients from a mini-batch of data points to approximate $\nabla F_k(w)$, and the aggregated mini-batch updates are applied on the server. \textit{Right: Local updating schemes.} Each device immediately applies local updates, e.g., gradients, after they are computed and a server performs a global aggregation after a variable number of local updates. Local-updating schemes can reduce communication by performing additional work locally.}
    \label{fig:local-updating} 
\end{figure}

Mini-batch optimization methods, which involve extending classical stochastic methods to process multiple data points at a time, have emerged as a popular paradigm for distributed machine learning in data center environments~\cite{dekel2012optimal, shalev2013accelerated, shamir2014distributed, qu2015quartz, richtarik2016distributed}. 
In practice, however, they have been shown to have limited flexibility to adapt to  communication-computation trade-offs that would maximally leverage distributed data processing~\cite{COCOA_Smith_2016,local_SGD_stich_18}.
In response, several recent methods have been proposed to improve communication-efficiency in distributed settings by allowing for a variable number of \textit{local updates} to be applied on each machine in parallel at each communication round, making the amount of computation versus communication substantially more flexible. 
For convex objectives, distributed local-updating \emph{primal-dual} methods have emerged as a popular way to tackle such a problem~\cite{COCOA_Smith_2016,lee2015distributed,ma2015adding,jaggi2014communication,yang2013trading}. These approaches leverage duality structure to effectively decompose the global objective into subproblems that can be solved in parallel at each communication round. Several distributed local-updating \emph{primal} methods have also been proposed, which have the added benefit of being applicable to non-convex objectives~\cite{elastic_SGD_zhang_LeCun_2015, AIDE_reddi_16}. These methods drastically improve performance in practice, and have been shown to achieve orders-of-magnitude speedups over traditional mini-batch methods or distributed approaches like ADMM~\cite{boyd2011distributed} in real-world data center environments. We provide an intuitive illustration of local updating methods in Figure \ref{fig:local-updating}.

In federated settings, optimization methods that allow for flexible local updating and low client participation have become the de facto solvers~\cite{mcmahan2016FedAvg, fed_multitask_smith_2017, sahu2018convergence}. 
The most commonly used method for federated learning is Federated Averaging (\fedavg)~\cite{mcmahan2016FedAvg}, a method based on averaging local stochastic gradient descent (SGD) updates for the primal problem. \fedavg has been shown to work well empirically, particularly for non-convex problems, but comes without convergence guarantees and can diverge in practical settings when data are heterogeneous~\cite{sahu2018convergence}. We discuss methods to handle such statistical heterogeneity in more detail in Section~\ref{related_work:non_iid_convergence}.

\subsubsection{Compression Schemes} 

While local updating methods can reduce the total \textit{number of communication rounds}, model compression schemes such as sparsification, subsampling, and quantization can significantly reduce the \textit{size of messages} communicated at each round. These methods have been extensively studied, both empirically and theoretically, in previous literature for distributed training in data center environments; we defer the readers to~\cite{wang2018atomo,zhang2016zipml} for a more complete review. In federated environments, the low participation of devices, non-identically distributed local data, and  local updating schemes pose novel challenges to these model compression approaches.
For instance, 
the commonly-used error compensation techniques in classical distributed learning~\cite{seide20141} cannot be directly extended to federated settings as the errors accumulated locally may be stale if the devices are not frequently sampled. 
Nevertheless, several works have provided practical strategies in federated settings, such as forcing the updating models to be sparse and low-rank; performing quantization with structured random rotations~\cite{konevcny2016federated}; using lossy compression and dropout to reduce server-to-device  communication~\cite{caldas2018expanding}; and applying Golomb lossless encoding~\cite{sattler2019robust}.  
From a theoretical perspective, while prior work has explored convergence guarantees with low-precision training in the presence of non-identically distributed data~\citep[e.g.,][]{tang2018communication},  the assumptions made do not take into consideration common characteristics of the federated setting, such as low device participation or locally-updating optimization methods. 

\begin{figure}[t]
    \centering
    \begin{subfigure}{0.3\textwidth}
    \includegraphics[width=\textwidth]{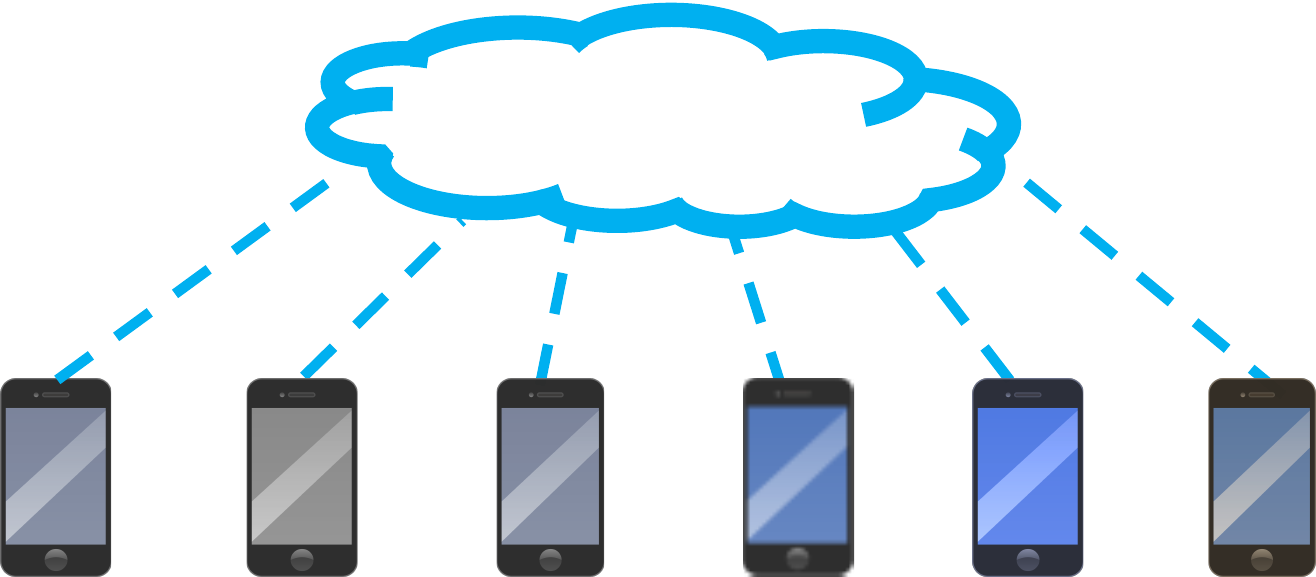}
    \end{subfigure}
    \hspace{0.14\textwidth}
    \begin{subfigure}{0.37\textwidth}
    \includegraphics[width=\textwidth]{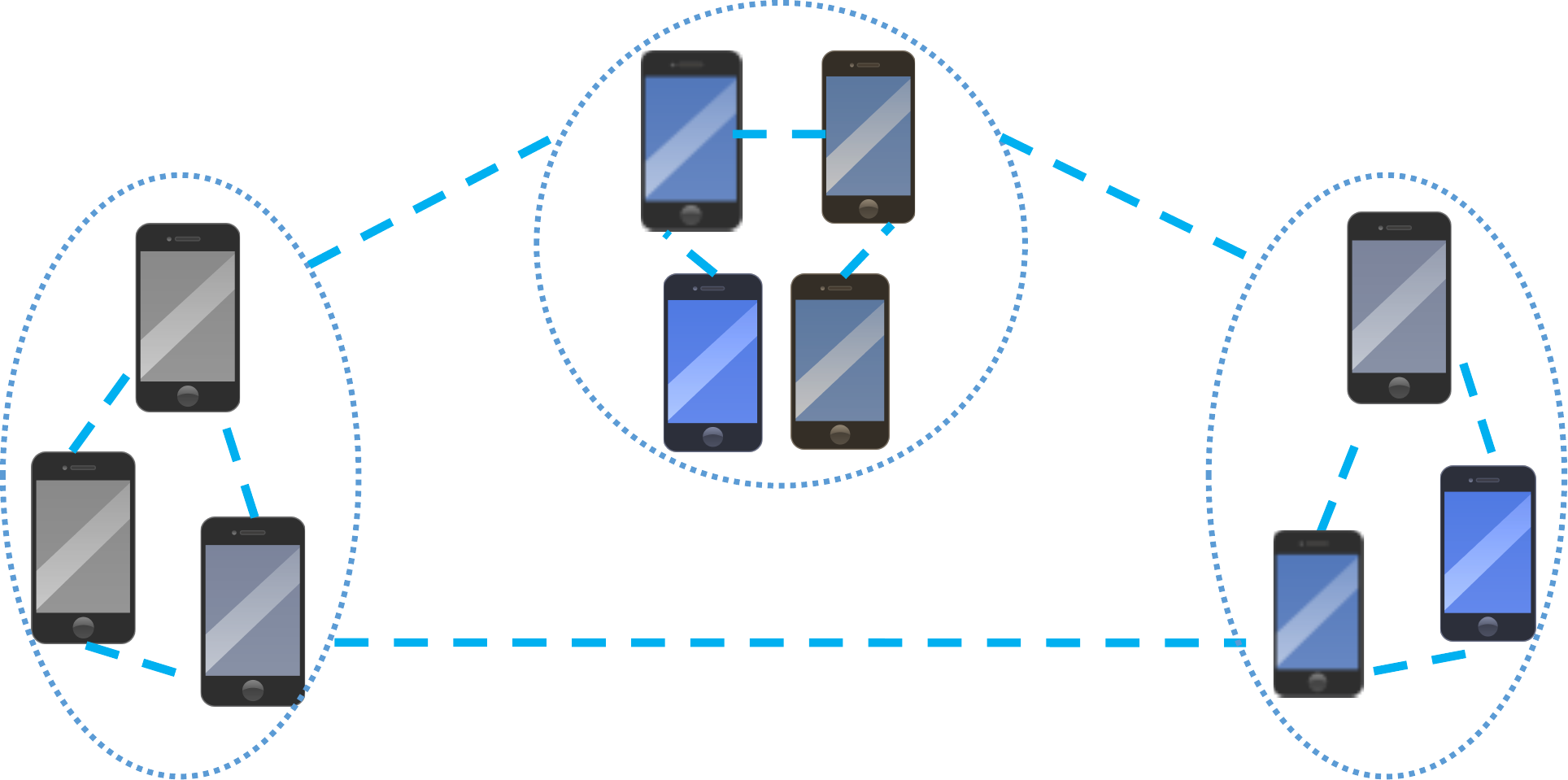}
    \end{subfigure}
    \caption{Centralized vs. decentralized topologies. In the typical federated learning setting and as a focus of this article, we assume a star network (left) where a server connects with all remote devices. Decentralized topologies (right) are a potential alternative when communication to the server becomes a bottleneck.}
    \label{fig:decentralized} 
\end{figure}

\subsubsection{Decentralized Training}\label{sec:related_work_communication_decentralized}
In federated learning, a star network (where a central server is connected to a network of devices, as in the left panel of Figure \ref{fig:decentralized}) is the predominant communication topology; we therefore  focus on the star-network setting in this article. However, we briefly discuss decentralized topologies (where devices only communicate with their neighbors, e.g., the right panel of Figure \ref{fig:decentralized}) as a potential alternative. In data center environments, decentralized training has been demonstrated to be faster than centralized training when operating on networks with low bandwidth or high latency; we defer readers to~\cite{he2018cola,lian2017can} for a more comprehensive review. 
Similarly, in federated learning, decentralized algorithms can in theory reduce the high communication cost on the central server. Some recent works~\cite{he2018cola,Lalitha2019decentralized} have investigated decentralized training over heterogeneous data with local updating schemes. However, they are either restricted to linear models~\cite{he2018cola} or assume full device participation~\cite{Lalitha2019decentralized}. 
Finally, hierarchical communication patterns have also been proposed~\cite{liu2019edge, lin2018don} to further ease the burden on the central server, by first leveraging \emph{edge servers} to aggregate the updates from edge devices and then relying on a \emph{cloud server} to aggregate updates from edge servers. While this is a promising approach to reduce communication, it is not applicable to all networks, as this type of physical hierarchy may not exist or be known a priori.

\subsection{Systems Heterogeneity}\label{sec:related_work_systems}

\begin{figure}[t]
    \centering
    \includegraphics[width=0.93\textwidth]{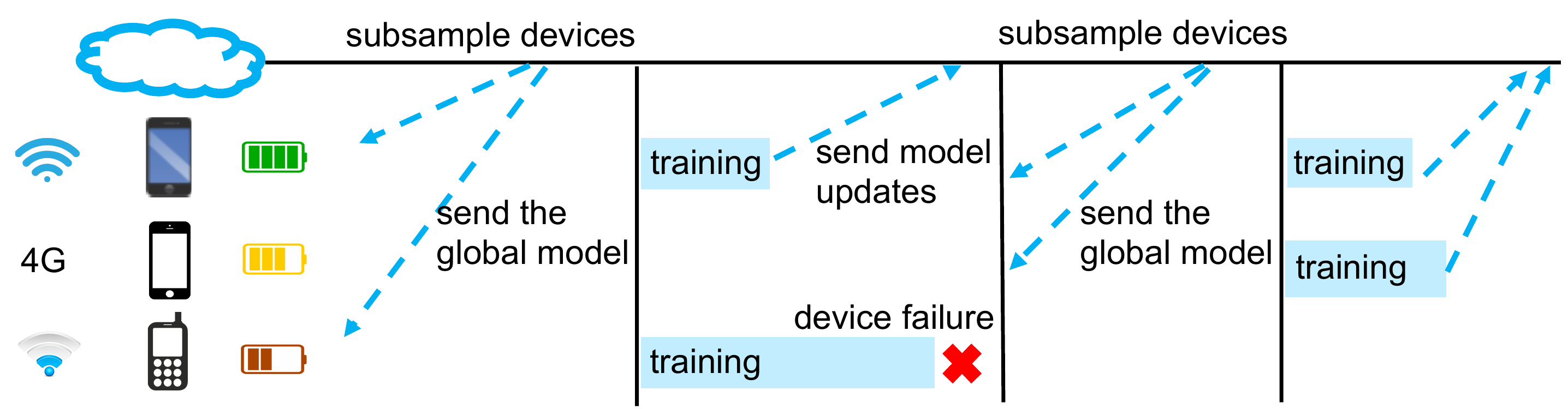}
    \caption{Systems heterogeneity in federated learning. Devices may vary in terms of network connection, power, and hardware.  Moreover, some of the devices may drop at any time during training. Therefore, federated training methods must tolerate heterogeneous systems environments and low participation of devices, i.e., they must allow for only a small subset of devices to be active at each round.} 
    \label{fig:sys_hetero}
\end{figure}

In federated settings, there is significant variability in the \emph{systems} characteristics across the network, as devices 
may differ in terms of hardware, network connectivity, and battery power. As depicted in Figure \ref{fig:sys_hetero}, these systems characteristics make issues such as stragglers significantly more prevalent than in typical data center environments. We roughly group several key directions to handle systems heterogeneity into: (i) asynchronous communication, (ii) active device sampling, and (ii) fault tolerance. As mentioned in  Section~\ref{sec:related_work_communication_decentralized}, we assume a star topology in our following discussions.

\subsubsection{Asynchronous Communication}\label{sec: related_work_system_commu} 

 In traditional data center settings, synchronous and asynchronous schemes are both commonly used to parallelize iterative optimization algorithms, with each approach having pros and cons. Synchronous schemes are simple and guarantee a serial-equivalent computational model, but they are also more susceptible to stragglers in the face of device variability. Asynchronous schemes are an attractive approach to mitigate stragglers in heterogeneous environments, particularly in shared-memory systems~\cite{recht2011hogwild, duchi2013estimation, zinkevich2010parallelized, ho2013more, dai2015high}. However, they typically rely on bounded-delay assumptions to control the degree of staleness, which for device $k$ depends on the number of other devices that have updated since device  $k$ pulled from the central server. While asynchronous parameter servers have been successful in distributed data centers~\citep[e.g.,][]{zinkevich2010parallelized, ho2013more, dai2015high}, classical bounded-delay assumptions can be unrealistic in federated settings,  where the delay may be on the order of hours to days, or completely unbounded.  
 
\subsubsection{Active Sampling}

In federated networks, typically only a small subset of devices participate at each round of training. However, the vast majority of federated 
methods, e.g. those described in~\cite{mcmahan2016FedAvg, fed_multitask_smith_2017, sahu2018convergence, bonawitz2019towards, he2018cola},
are \emph{passive} in that they do not aim to influence which devices participate. An alternative approach involves \emph{actively} selecting participating devices at each round. 
For example, \citet{nishio2018client} explore novel device sampling policies based on systems resources, with the aim being for the server to aggregate as many device updates as possible within a pre-defined time window. 
Similarly, \citet{kang2019incentive} take into account systems overheads incurred on each device when designing incentive mechanisms to encourage devices with higher-quality data to participate in the learning process. However, these methods assume a static model of the systems characteristics of the network; it remains open how to extend these approaches to handle \textit{real-time}, device-specific fluctuations in computation and communication delays. Moreover, while these methods primarily focus on systems variability to perform active sampling, we note that it is also worth considering actively sampling a set of small but sufficiently representative devices based on the underlying \emph{statistical} structure. 

\subsubsection{Fault Tolerance} 
Fault tolerance has been extensively studied in the  systems community and is a fundamental consideration of classical distributed systems~\cite{tanenbaum2007distributed,liu2013data,castro1999practical}. Recent works have also investigated fault tolerance specifically for machine learning workloads in data center environments~\cite[e.g.,][]{tang2018distributed,pmlr-v97-qiao19a}. 
When learning over remote devices, however, fault tolerance becomes more critical as it is common for some participating devices to drop out at some point before the completion of the given training iteration~\cite{bonawitz2019towards}. One practical strategy is to simply ignore such device failure~\cite{bonawitz2019towards}, 
which may introduce  bias into the device sampling scheme if the failed devices have specific data characteristics. For instance, devices from remote areas may be more likely to drop due to poor network connections and thus the trained federated model will be biased towards  devices with favorable network conditions. Theoretically, while several recent works have investigated convergence guarantees of variants of federated learning methods~\cite{yu2019linear,wang2018adaptive,jiang2018linear,yu2018parallel}, few analyses allow for low participation~\cite[e.g.,][]{sahu2018convergence, fed_multitask_smith_2017}, or study directly the effect of dropped devices.

\emph{Coded computation} is another option to tolerate device failures by introducing algorithmic redundancy.  Recent works have explored using codes to speed up distributed machine learning training~\citep[e.g.,][]{lee2017speeding, tandon2017gradient,Charles2018GradientCU,reisizadeh2019coded,carles2017approx}. For instance, in the presence of stragglers, gradient coding and its variants~\cite{tandon2017gradient,Charles2018GradientCU,carles2017approx} carefully replicate data blocks (as well as the gradient computation on those data blocks) across computing nodes to obtain either exact or inexact recovery of the true gradients.
While this is a seemingly promising approach for the federated setting, these methods face fundamental challenges in federated networks as sharing data/replication across devices is often infeasible due to privacy constraints and the scale of the network. 

\subsection{Statistical Heterogeneity}\label{sec:related_work_statistical}

\begin{figure}
    \centering
    \begin{subfigure}{0.22\textwidth}
    \includegraphics[width=\textwidth]{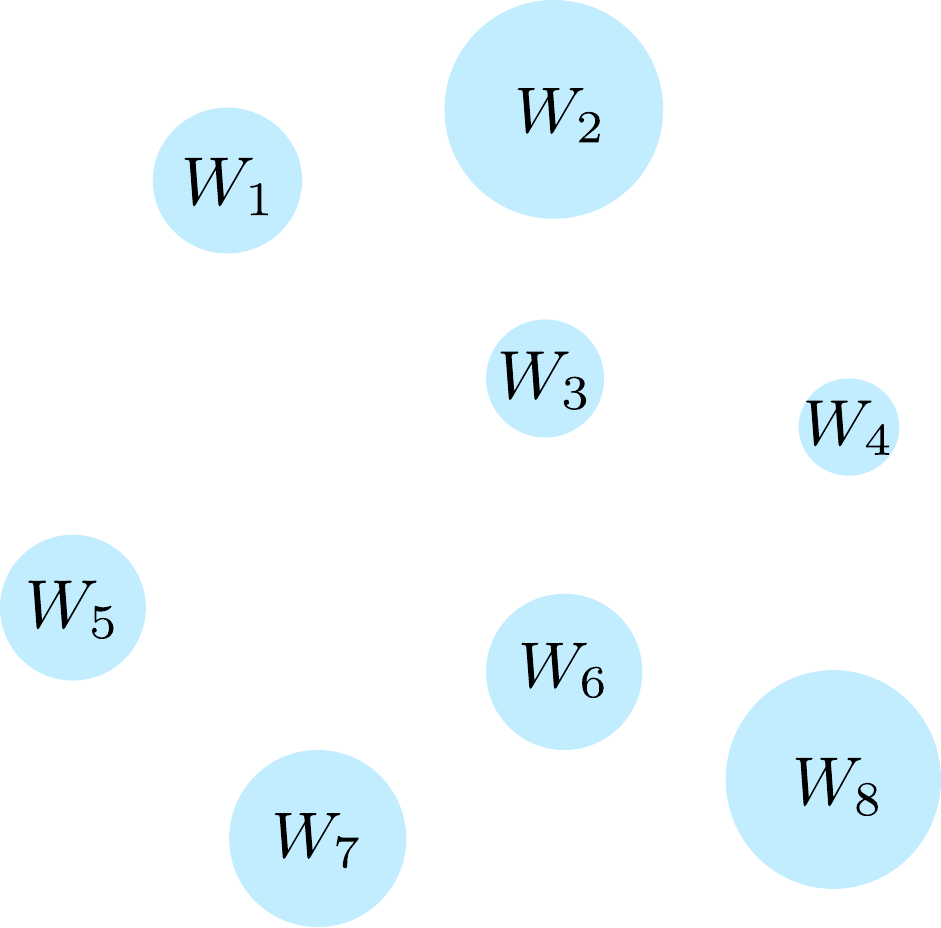}
    \subcaption{Learn personalized models for each device; do not learn from peers.}
    \end{subfigure}
    \hfill
    \begin{subfigure}{0.22\textwidth}
    \includegraphics[width=\textwidth]{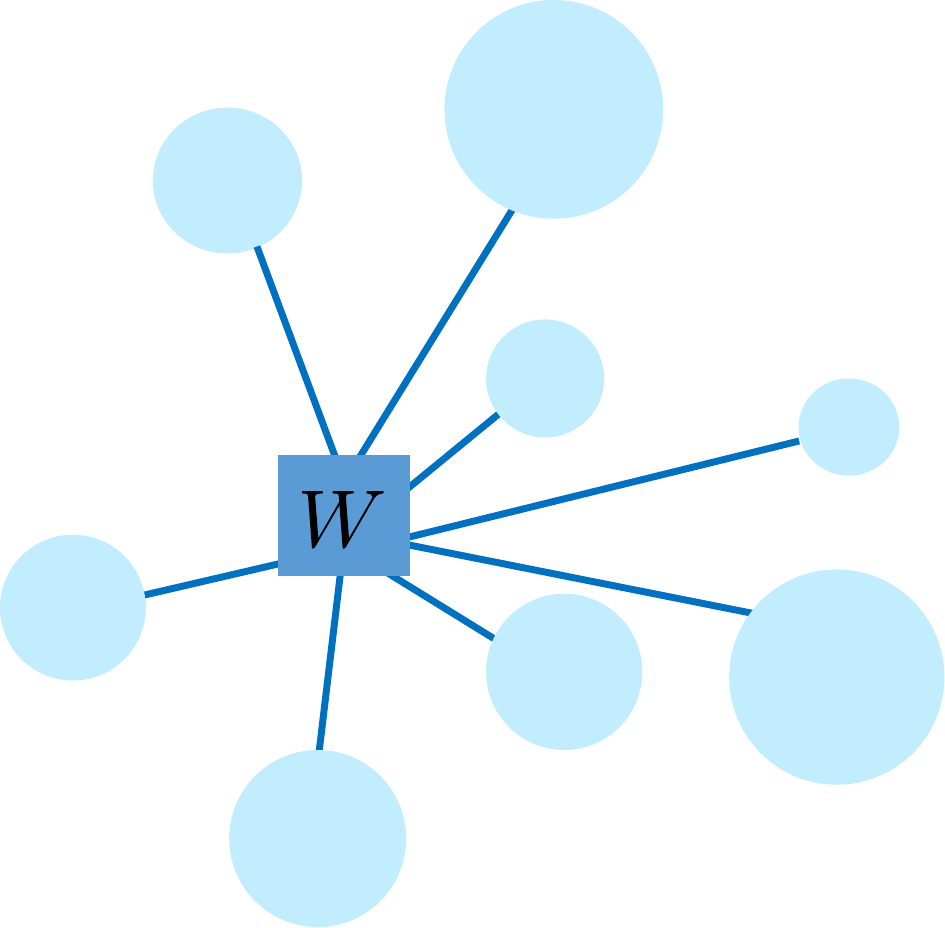}
    \subcaption{Learn a global model; learn from peers.}
    \end{subfigure}
    \hfill
    \begin{subfigure}{0.22\textwidth}
    \includegraphics[width=\textwidth]{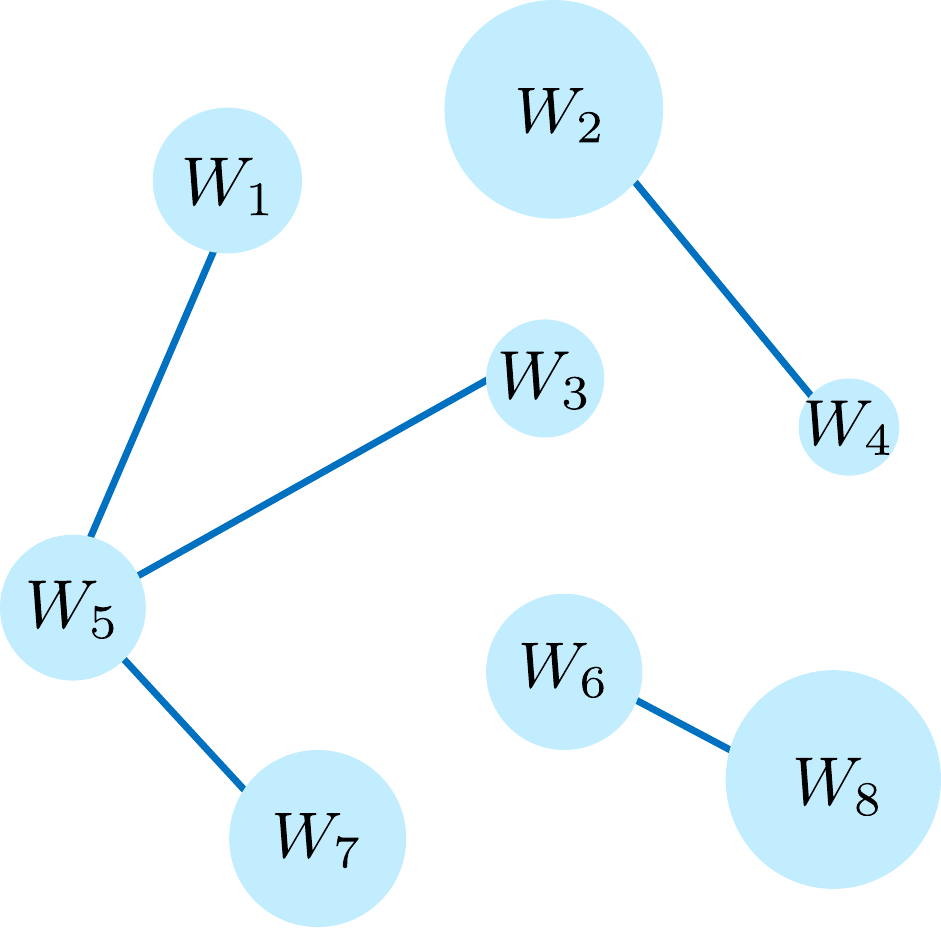}
    \subcaption{Learn personalized models for each device; learn from peers.}
    \end{subfigure}
    \caption{Different  modeling approaches in federated networks. Depending on properties of the data, network, and application of interest, one may choose to (a) learn separate models for each device, (b) fit a single global model to all devices, or (c) learn related but distinct models in the network.}
    \label{fig:data_model}
\end{figure}

Challenges arise when training federated models from data that is not identically distributed across devices, both in terms of modeling the data (as depicted in Figure~\ref{fig:data_model}), and in terms of analyzing the convergence behavior of associated training procedures. We discuss related work in these directions below.

\subsubsection{Modeling Heterogeneous Data}
There exists a large body of literature in machine learning that has modeled statistical heterogeneity via methods such as meta-learning~\cite{thrun2012learning} and multi-task learning~\cite{Caruana:1998ml,evgeniou2004regularized}; these ideas have been recently extended to the federated setting~\cite{chen2018federated, fed_multitask_smith_2017,varia_mtl,khodak2019adaptive,eichner2019semi,zhao2018federated}.  For instance, \mocha~\cite{fed_multitask_smith_2017}, an optimization framework designed for the federated setting, can allow for  personalization  by learning \emph{separate} but related models for each device while leveraging a shared representation via multi-task learning.  This method has provable theoretical convergence guarantees for the considered objectives, but is limited in its ability to scale to massive networks and is restricted to convex objectives. Another approach~\cite{varia_mtl} models the star topology as a Bayesian network and performs variational inference during learning. Although this method can handle non-convex models, it is expensive to generalize to large federated networks. \citet{khodak2019adaptive} provably meta-learn a within-task learning rate using multi-task information (where each task corresponds to a device) and have demonstrated improved empirical performance over vanilla \fedavg.  
\citet{eichner2019semi} investigate a pluralistic solution (adaptively choosing between a global model and device-specific models) to address the cyclic patterns in data samples during federated training. \citet{zhao2018federated} explore transfer learning for personalization by running \fedavg after training a global model centrally on some shared proxy data. Despite these recent advances, key challenges still remain in making methods for heterogeneous modeling that are robust, scalable, and automated in federated settings.

 When modeling federated data, it may also be important to consider issues beyond accuracy, such as \textit{fairness}. 
 In particular, naively solving an aggregate loss function such as in \eqref{eq:original_obj} may implicitly advantage or disadvantage some of the devices, as the learned model may become biased towards devices with larger amounts of data, or (if weighting devices equally), to commonly occurring groups of devices. Recent works have proposed modified modeling approaches that aim to reduce the variance of the model performance across devices. Some heuristics simply perform a varied number of local updates based on local loss~\cite{huang2018loadaboost}. Other more principled approaches include Agnostic Federated Learning~\cite{google2019agnostic}, which optimizes the centralized model for any target distribution formed by a mixture of the client distributions via a minimax optimization scheme. 
 Another more general approach is taken by \citet{li2019fair}, which proposes an objective called $q$-FFL in which devices with higher loss are given higher relative weight to encourage less variance in the final accuracy distribution. Beyond issues of fairness, we note that aspects such as accountability and interpretability in federated learning are additionally worth exploring, but may be challenging due to the scale and heterogeneity of the network.

\subsubsection{Convergence Guarantees for Non-IID Data}\label{related_work:non_iid_convergence}
 
Statistical heterogeneity also presents novel challenges in terms of analyzing the convergence behavior in federated settings---even when learning a single global model.
Indeed, when data is not identically distributed across devices in the network, methods such as \fedavg have been shown to diverge in practice~\cite{sahu2018convergence, mcmahan2016FedAvg}. Parallel SGD and related variants, which make local updates similar to \fedavg, have been analyzed in the I.I.D. setting~\cite{lin2018don,AIDE_reddi_16,shamir2014communication,local_SGD_stich_18,cooperative_SGD_Joshi_18,parallel_SGD_Srebro_18,elastic_SGD_zhang_LeCun_2015,zhou2017convergence,wang2018giant,jianyu2018adaptive}. 
However, the results rely on the premise that each local
solver is a copy of the same stochastic process (due to the I.I.D. assumption), which is not the case in typical federated settings. 
To understand the performance of \fedavg in statistically heterogeneous settings, \fedprox~\cite{sahu2018convergence} has recently been proposed. \fedprox makes a small modification to the \fedavg method to help ensure convergence, both theoretically and in practice. \fedprox can also be interpreted as a generalized, reparameterized version of \fedavg that has  practical ramifications in the context of accounting for systems heterogeneity across devices.
Several other works~\cite{wang2018adaptive,yu2019linear,jiang2018linear,yu2018parallel} have also explored convergence guarantees in the presence of heterogeneous data with different assumptions, e.g., convexity~\cite{wang2018adaptive} or uniformly bounded gradients~\cite{yu2018parallel}. 
There are also heuristic approaches that aim to tackle statistical heterogeneity, either by sharing local device data or some server-side proxy data~\cite{huang2018loadaboost,jeong2018communication,zhao2018federated}. However, these methods may be unrealistic: in addition to imposing burdens on network bandwidth, sending local data to the server~\cite{jeong2018communication} violates the key privacy assumption of federated learning, and sending globally-shared proxy data to all devices~\cite{huang2018loadaboost,zhao2018federated} requires effort to carefully generate or collect such auxiliary data.

\subsection{Privacy}

Privacy concerns often motivate the need to keep raw data on each device local in federated settings. However, sharing other information such as model updates as part of the training process can also leak sensitive user information~\cite{bhowmick2018protection,carlini2018secret,melis2018inference,fredrikson2015model}. For instance, \citet{carlini2018secret} demonstrate that one can extract sensitive text patterns, e.g., a specific credit card number, from a recurrent neural network trained on users' language data. 
Given increasing interest in privacy-preserving learning approaches, in Section~\ref{sec:related_work_privacy_ml}, we first briefly revisit prior work on enhancing privacy in the general (distributed) machine learning setting.  We then review recent privacy-preserving methods specifically designed for federated settings in Section~\ref{sec:related_work_privacy_fl}.

\subsubsection{Privacy in Machine Learning}\label{sec:related_work_privacy_ml}

Privacy-preserving learning  has been extensively studied by the machine learning~\citep[e.g.,][]{mcmahan2018diff}, systems~\citep[e.g.,][]{bonawitz2019towards,agrawal2000privacy}, and theory~\citep[e.g.,][]{feldman2018privacy,Lindell2000privacy} communities. Three main strategies, each of which we will briefly review, include differential privacy to communicate noisy data sketches, homomorphic encryption to operate on encrypted data, and secure function evaluation or multiparty computation.  

Among these various privacy approaches, \textit{differential privacy}~\cite{dwork2006calibrating,dwork2014algorithmic,dwork2011firm} is most widely used due to its strong information theoretic guarantees, algorithmic simplicity, and relatively small systems overhead. 
Simply put, a randomized mechanism is differentially private if the change of one input element will not result in too much difference in the output distribution; this means that one cannot draw any conclusions about whether or not a specific sample is used in the learning process. 
Such sample-level privacy can be achieved in many learning tasks~\cite{chaudhuri2011differentially,bassily2014private,abadi2016deep,papernot2016semi,papernot2018scalable,iyengar2019towards}.  
For gradient-based learning methods, a popular approach is to apply differential privacy by randomly perturbing the intermediate output at each iteration~\citep[e.g.,][]{abadi2016deep,bassily2014private,wu2017bolt}. Before applying the perturbation, e.g., via Gaussian noise~\cite{abadi2016deep}, Laplacian noise~\cite{MelisDC16}, or Binomial noise~\cite{agarwal2018cpsgd}, it is common to clip the gradients in order to bound the influence of each example on the overall update. There exists an inherent trade-off between  differential privacy and  model accuracy, as adding more noise results in greater privacy, but may compromise accuracy significantly.
Despite the fact that differential privacy is the de facto metric for privacy in machine learning, 
there are many other privacy definitions, 
such as $k$-anonymity~\cite{el2008protecting}, $\delta$-presence~\cite{4912209} and distance correlation~\cite{vepakomma2019reducing}, that may be applicable to different learning problems~\cite{wagner2018technical}.

Beyond differential privacy, homomorphic encryption can be used to secure the learning process by computing on encrypted data, although it has currently been applied in limited settings, e.g.,  training linear models~\cite{nikolaenko2013privacy} or involving only a few entities~\cite{yuan2013privacy}. When the sensitive datasets are distributed across different data owners, another natural option is to perform privacy-preserving learning via secure function evaluation (SFE) or secure multiparty computation (SMC). The resulting protocols can enable multiple parties to collaboratively compute an agreed-upon function without leaking input information from any party except for what can be inferred from the output~\citep[e.g.,][]{goryczka2015comprehensive,chaum1988dining,riazi2018chameleon}. Thus, while SMC  cannot guarantee protection from information leakage,  it can be combined with differential privacy to achieve stronger privacy guarantees. However, approaches along these lines may not be applicable to large-scale machine learning scenarios as they incur substantial additional communication and computation costs. 
Moreover,  SMC protocols need to be carefully designed and implemented for each operation in the targeted learning algorithm~\cite{chen2019secure,mohassel2018aby}.
We defer interested readers to~\cite{bost2015machine,Rouhani2018design} for a more comprehensive review of the approaches based on homomorphic encryption and SMC. 

\begin{figure}
    \centering
    \begin{subfigure}{0.28\textwidth}
    \includegraphics[width=\textwidth]{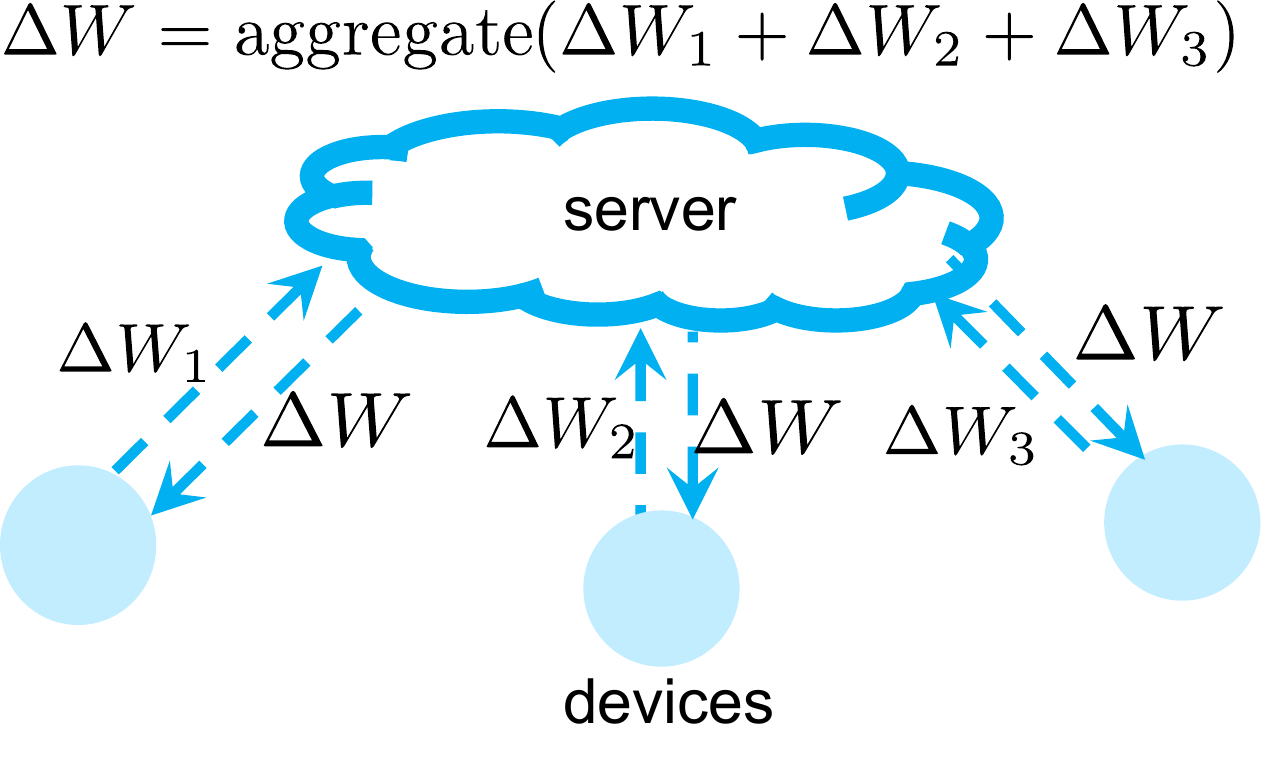}
    \subcaption{Federated learning without additional privacy protection mechanisms.}
    \end{subfigure}
    \hfill
    \begin{subfigure}{0.31\textwidth}
    \includegraphics[width=\textwidth]{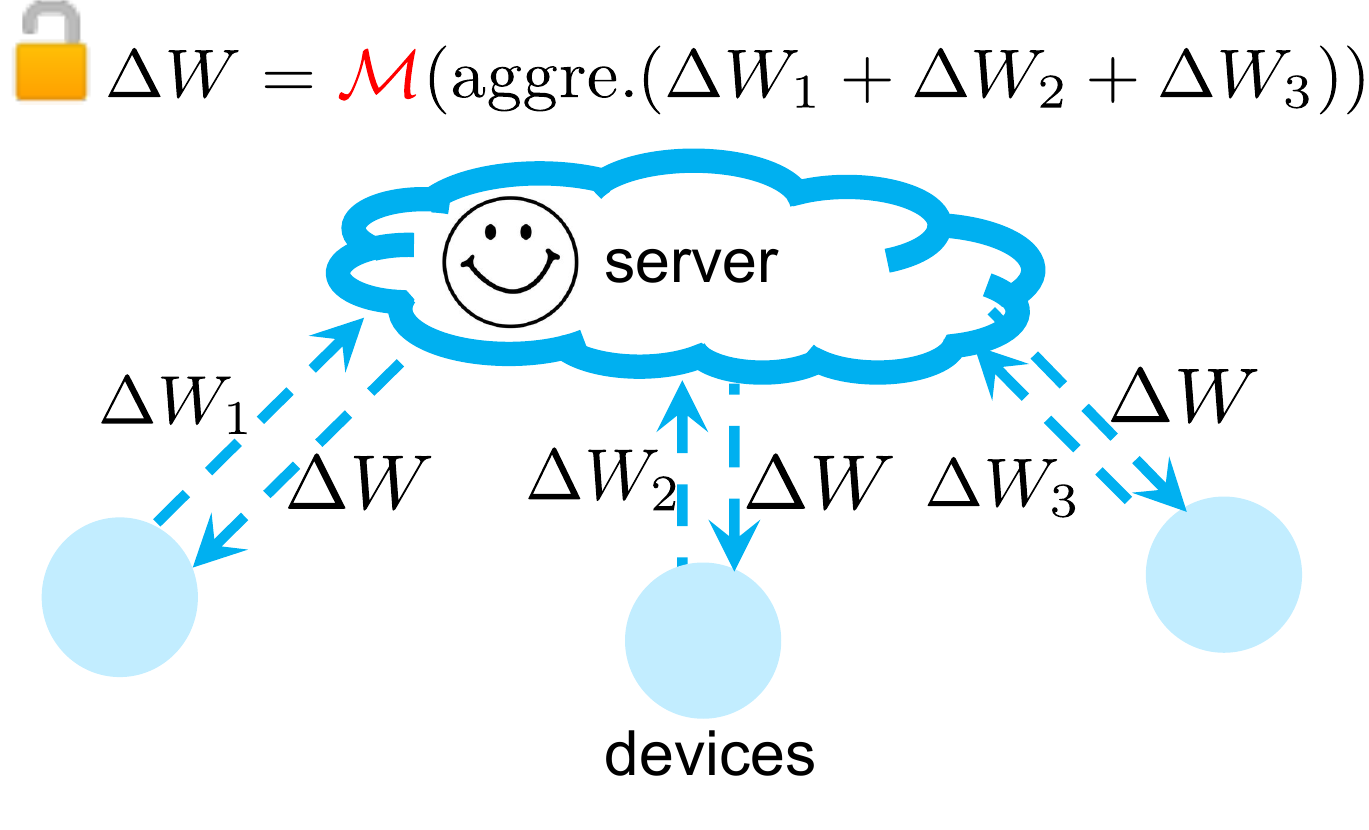}
    \subcaption{Global privacy, where a trusted server is assumed.}
    \end{subfigure}
    \hfill
    \begin{subfigure}{0.35\textwidth}
    \includegraphics[width=\textwidth]{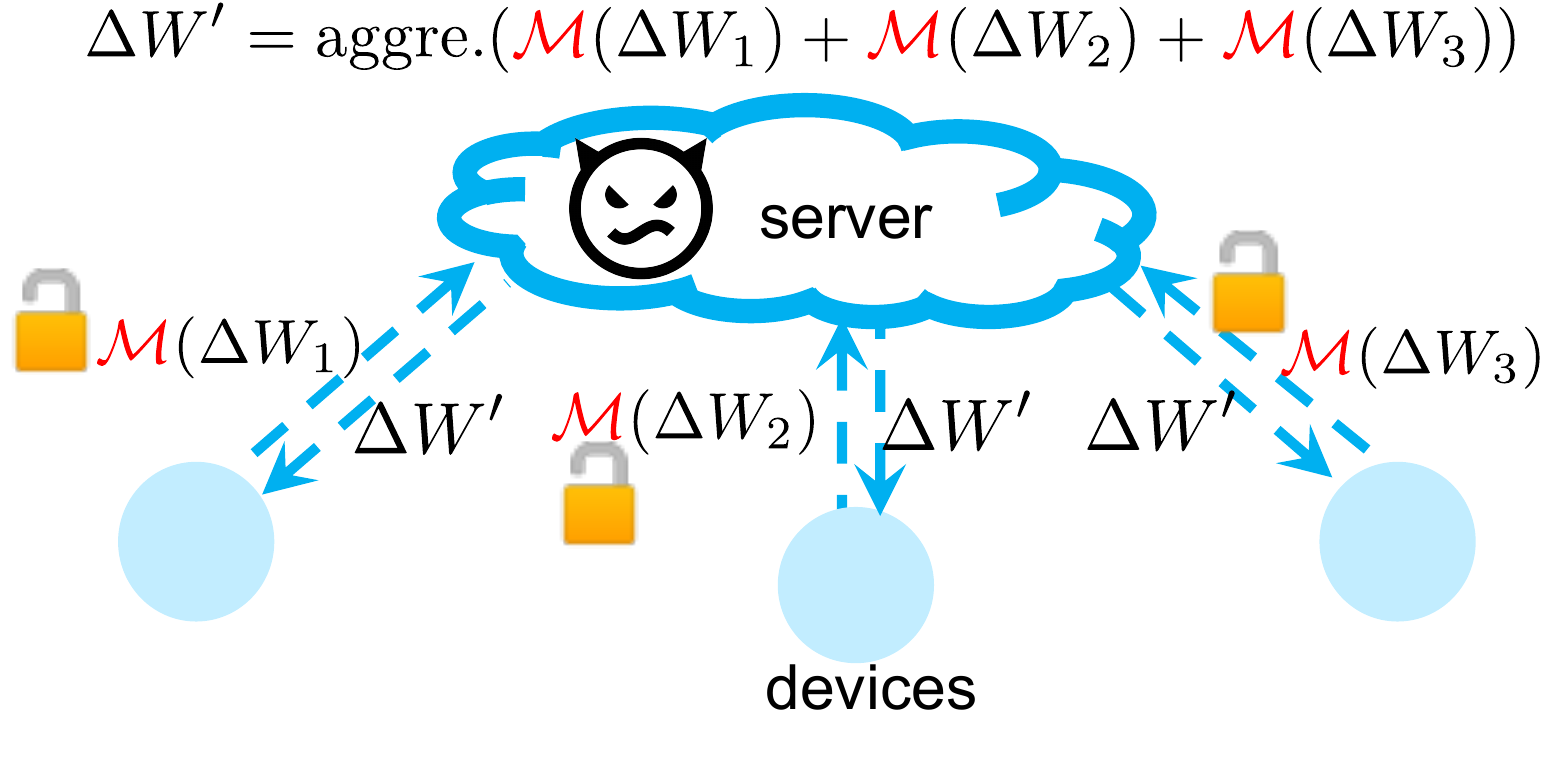}
    \subcaption{Local privacy, where the central server might be malicious.}
    \end{subfigure}
    \caption{An  illustration of different privacy-enhancing mechanisms in one  round of federated learning. $\mathcal{M}$ denotes a randomized mechanism used to privatize the data. With global privacy (b), the model updates are private to all third parties other than a single trusted party (the central server). With local privacy (c), the individual model updates are also private to the server.}
    \label{fig:privacy_model}
\end{figure}

\subsubsection{Privacy in Federated Learning}
\label{sec:related_work_privacy_fl}

The federated setting poses novel challenges to existing privacy-preserving algorithms. Beyond providing rigorous privacy guarantees, it is necessary to develop methods that are computationally cheap, communication-efficient, and tolerant to dropped devices---all without overly compromising accuracy. 
Although there are a variety of privacy definitions in federated learning~\cite{bhowmick2018protection, carlini2018secret,mcmahan2018diff,thakkar2019differentially,geyer2017differentially,li_gbml19}, typically they can be classified into two categories: \textit{global privacy}  and \textit{local privacy}.  As demonstrated in Figure~\ref{fig:privacy_model},
global privacy requires that the model updates generated at each round are private to all untrusted third parties other than the central server, while local privacy further requires that the updates are also private to the server. 

Current works that aim to improve the privacy of federated learning typically build upon previous classical cryptographic protocols such as SMC~\cite{bonawitz2017practical,Ghazi2019scalable} and differential privacy~\cite{geyer2017differentially, mcmahan2018diff,bhowmick2018protection,agarwal2018cpsgd}. \citet{bonawitz2017practical} introduce an SMC protocol to protect individual model updates. The central server is not able to see any local updates, but can still observe the exact aggregated results at each round. SMC is a lossless method, and can retain the original accuracy with a very high privacy guarantee. 
However, the resulting method incurs significant extra communication cost. Other works~\cite{mcmahan2018diff, geyer2017differentially} apply differential privacy to federated learning and offer global differential privacy. These approaches have a number of hyperparameters that affect communication and accuracy that must be carefully chosen, though follow up work~\cite{thakkar2019differentially} proposes adaptive gradient clipping strategies to help alleviate this issue. In the case where stronger privacy guarantees are required, \citet{bhowmick2018protection} introduce a  relaxed version of local privacy by limiting  the power of potential adversaries. It affords stronger privacy guarantees than global privacy, and has better model performance than strict local privacy. \citet{li_gbml19} propose locally differentially-private algorithms in the context of meta-learning, which can be applied to federated learning with personalization, while also
providing provable learning guarantees in convex settings.
In addition, differential privacy can  be combined with model compression techniques to reduce communication and obtain privacy benefits simultaneously~\cite{agarwal2018cpsgd}.

\section{Future Directions}
\label{sec: future_work}
Federated learning is an active and ongoing area of research. Although recent work has begun to address the challenges discussed in Section~\ref{sec: related_work}, there are a number of critical open directions yet to be explored.
In this section, we briefly outline a few promising research directions surrounding the previously discussed challenges (expensive communication, systems heterogeneity, statistical heterogeneity, and privacy concerns), and introduce additional challenges regarding issues such as productionizing and benchmarking in federated settings.
\begin{itemize}
    \item \textbf{Extreme communication schemes.} It remains to be seen how much communication is necessary in federated learning. Indeed, it is well-known that optimization methods for machine learning can tolerate a lack of precision; this error can in fact help with generalization~\cite{yao2007early}. While one-shot or divide-and-conquer communication schemes have been explored in  traditional data center settings~\cite{zhangduchi,DFC}, the behavior of these methods is not well-understood in massive or statistical heterogeneous networks. Similarly, one-shot/few-shot heuristics~\cite{yurochkin2019bayesian, Neel_GEMS,Neel_one_shot} have recently been proposed for the federated setting, but have yet to be theoretically analyzed or  evaluated at scale.
    \item \textbf{Communication reduction and the Pareto frontier.} We discussed  several ways to reduce communication in federated training, such as local updating and model compression. In order to create a realistic system for federated learning, it is important to understand how these techniques \textit{compose} with one another, and to \textit{systematically} analyze the trade-off  between accuracy and communication for each approach. In particular, the most useful techniques will demonstrate improvements at the Pareto frontier---achieving an accuracy greater than any other approach under the same communication budget, and ideally, across a wide range of communication/accuracy profiles.
    Similar comprehensive analyses have been performed for efficient neural network inference~\cite[e.g.,][]{bolukbasi2017adaptive}, and are necessary in order to compare communication-reduction techniques for federated learning  in a meaningful way.
    \item  \textbf{Novel models of asynchrony.} As discussed in Section~\ref{sec: related_work_system_commu}, two communication schemes most commonly studied in distributed optimization are bulk synchronous approaches and asynchronous approaches (where it is assumed that the delay is bounded). These schemes are more realistic in  data center settings---where worker nodes are typically \emph{dedicated} to the workload, i.e., they are ready to `pull' their next job from the central node immediately after they `push' the results of their previous job. In contrast, in federated networks, each device is often \emph{undedicated} to the task at hand and most devices are not active on any given iteration. 
    Therefore, it is worth studying the effects of this more realistic \emph{device-centric} communication scheme---in which each device can decide when to `wake up' and interact with the central server in an event-triggered manner.
    \item \textbf{Heterogeneity diagnostics.} 
    Recent works have aimed to quantify statistical heterogeneity through metrics such as local dissimilarity (as defined in the context of federated learning in~\cite{sahu2018convergence} and used for other purposes in works such as~\cite{yin2018gradient,SGD_under_growth_schmidt_2013,vaswani2018fast}) and earth mover's distance~\cite{zhao2018federated}. However, these metrics cannot be easily calculated over the federated network before training occurs. 
    The importance of these metrics motivates the following open questions: (i) Do simple diagnostics exist to quickly determine the level of heterogeneity in federated networks \textit{a priori}? (ii) Can analogous diagnostics be developed to quantify the amount of \emph{systems-related} heterogeneity? (iii) Can current or new definitions of heterogeneity be exploited to further improve the convergence of federated optimization methods? 
    \item \textbf{Granular privacy constraints.} The definitions of privacy outlined in Section~\ref{sec:related_work_privacy_fl} cover privacy at a local or global level with respect to all devices in the network. However, in practice, it may be necessary to define privacy on a more granular level, as privacy constraints may differ across devices or even across data points on a single device. For instance, \citet{li_gbml19} recently proposed sample-specific (as opposed to user-specific) privacy guarantees, thus providing a weaker form of privacy in exchange for more accurate models.  Developing methods to handle mixed (device-specific or sample-specific) privacy restrictions is an interesting and ongoing direction of future work.
    \item \textbf{Beyond supervised learning.} It is important to note that the methods discussed thus far have been developed with the task of \textit{supervised learning} in mind, i.e., they assume that labels exist for all of the data in the federated network. In practice, much of the data generated in realistic federated networks may be unlabeled or weakly labeled. Furthermore, the problem at hand may not be to fit a model to data as presented in~\eqref{eq:original_obj}, but instead to perform some exploratory data analysis, determine aggregate statistics, or run a more complex task such as reinforcement learning. Tackling problems beyond supervised learning in federated networks will likely require addressing similar challenges of scalability, heterogeneity, and privacy. 
    \item \textbf{Productionizing federated learning.} Beyond the major challenges discussed in this article, there are a number of practical concerns that arise when running federated learning in production. In particular, issues such as concept drift (when the underlying data-generation model changes over time); diurnal variations (when the devices exhibit different behavior at different times of the day or week)~\cite{eichner2019semi}; and cold start problems (when new devices enter the network) must be handled with care. We defer the readers to~\cite{bonawitz2019towards}, which discusses some of the practical systems-related issues that exist in production federated learning systems.
    \item \textbf{Benchmarks.} Finally, as federated learning is a nascent field, we are at a pivotal time to shape the developments made in this area and ensure that they are grounded in real-world settings, assumptions, and datasets. It is critical for the broader research communities to further build upon existing implementations and benchmarking tools, such as LEAF~\cite{caldas2018leaf} and TensorFlow Federated~\cite{tf_federated}, to facilitate both the reproducibility of empirical results and the dissemination of new solutions for federated learning.
\end{itemize}

\section{Conclusion}

In this article, we have provided an overview of federated learning, a learning paradigm where statistical models are  trained at the edge in distributed networks. We have discussed the unique properties and associated challenges of federated learning compared with traditional distributed data center computing and classical privacy-preserving learning. We  provided an extensive survey on classical results 
as well as more recent work specifically focused on federated settings. Finally, we have outlined out a handful of  open problems worth future research effort. Providing solutions to these problems will require interdisciplinary effort from a broad set of research communities.

\paragraph{Acknowledgement.} We thank Jeffrey Li and Mikhail Khodak for helpful discussions and comments. This work was supported in part by DARPA FA875017C0141, the National Science Foundation
grants IIS1705121 and IIS1838017, an Okawa Grant, a Google Faculty Award, an Amazon Web
Services Award,  a JP Morgan A.I. Research Faculty Award, a Carnegie Bosch Institute Research Award and the CONIX Research Center, one of
six centers in JUMP, a Semiconductor Research Corporation (SRC) program sponsored by DARPA.
Any opinions, findings, and conclusions or recommendations expressed in this material are those of
the author(s) and do not necessarily reflect the views of DARPA, the National Science Foundation, or
any other funding agency.

\bibliographystyle{abbrvnat}

{\small
\bibliography{ref}}

\end{document}